\documentclass[letterpaper, 10 pt, conference]{ieeeconf}  

\IEEEoverridecommandlockouts                              

\usepackage{cite}
\usepackage{amsmath,amssymb,amsfonts}
\usepackage{graphicx}
\usepackage{textcomp}
\usepackage{xcolor}
\usepackage{hyperref}[colorlinks,linkcolor=blue] 
\usepackage{amsmath, bm} 
\usepackage{algorithm}
\usepackage{algorithmicx}
\usepackage{algpseudocode}
\usepackage{threeparttable}
\usepackage{booktabs}
\usepackage{multirow}
\usepackage{marvosym} 

\title{\LARGE \bf
SSC3OD: Sparsely Supervised Collaborative 3D Object Detection \\from LiDAR Point Clouds\\
\thanks{\Letter: Corresponding author.}
\thanks{This work was supported by the National Natural Science Foundation of China under Grant U1934220 and U2268203.}
}

\author{Yushan Han\textsuperscript{1,2}, Hui Zhang\textsuperscript{1,2,\Letter}, Honglei Zhang\textsuperscript{1,2} and Yidong Li\textsuperscript{1,2}\\
\textit{1. Key Laboratory of Big Data \& Artificial Intelligence in Transportation, Ministry of Education} \\
\textit{2. School of Computer and Information Technology, Beijing Jiaotong University}\\
Beijing, China\\
\textit{E-mail:\{yushanhan, huizhang1, honglei.zhang, ydli\}@bjtu.edu.cn}
}

\begin{document}

\maketitle
\thispagestyle{empty}
\pagestyle{empty}

\begin{abstract}

Collaborative 3D object detection, with its improved interaction advantage among multiple agents, has been widely explored in autonomous driving. However, existing collaborative 3D object detectors in a fully supervised paradigm heavily rely on large-scale annotated 3D bounding boxes, which is labor-intensive and time-consuming. To tackle this issue, we propose a sparsely supervised collaborative 3D object detection framework SSC3OD, which only requires each agent to randomly label one object in the scene. Specifically, this model consists of two novel components, i.e., the pillar-based masked autoencoder (Pillar-MAE) and the instance mining module. The Pillar-MAE module aims to reason over high-level semantics in a self-supervised manner, and the instance mining module generates high-quality pseudo labels for collaborative detectors online. By introducing these simple yet effective mechanisms, the proposed SSC3OD can alleviate the adverse impacts of incomplete annotations. We generate sparse labels based on collaborative perception datasets to evaluate our method. Extensive experiments on three large-scale datasets reveal that our proposed SSC3OD can effectively improve the performance of sparsely supervised collaborative 3D object detectors.

\end{abstract}

\section{Introduction}
Collaborative perception \cite{caillot2022survey} plays an essential role in expanding the perspective of autonomous vehicles by leveraging the interactions among multiple agents \cite{OPV2V,where2comm,luo2020multiagent}, which has gained considerable attention both in academia and industry \cite{cao2022future,chen2022milestones,hu2022driver,hu2022review,wu2022uncertainty,huang2022survey,wang2022verification,zhang2022learning,henning2022situation}.
Recent pioneering works have been dedicated to the development of high-quality datasets \cite{DAIR-V2X, OPV2V, V2X-Sim, Xu2023V2V4Real}, effective fusion strategies \cite{fcooper,v2vnet,DiscoNet}, and communication mechanisms \cite{where2comm}. Besides, another research line explores how to overcome real-world issues such as latency \cite{SyncNet} and pose errors \cite{coalign}.
Despite great success, these studies on collaborative 3D object detection have adopted a fully supervised learning approach, which heavily relies on large-scale annotated 3D bounding boxes. However, collecting accurate annotations is labor-intensive and time-consuming, particularly for collaborative 3D object detection involving multiple agents. Thus, developing collaborative 3D object detectors that rely on lightweight object annotations is a significant issue in practical applications.

\begin{figure}[htbp]
    \centerline{\includegraphics[width=1.0\linewidth]{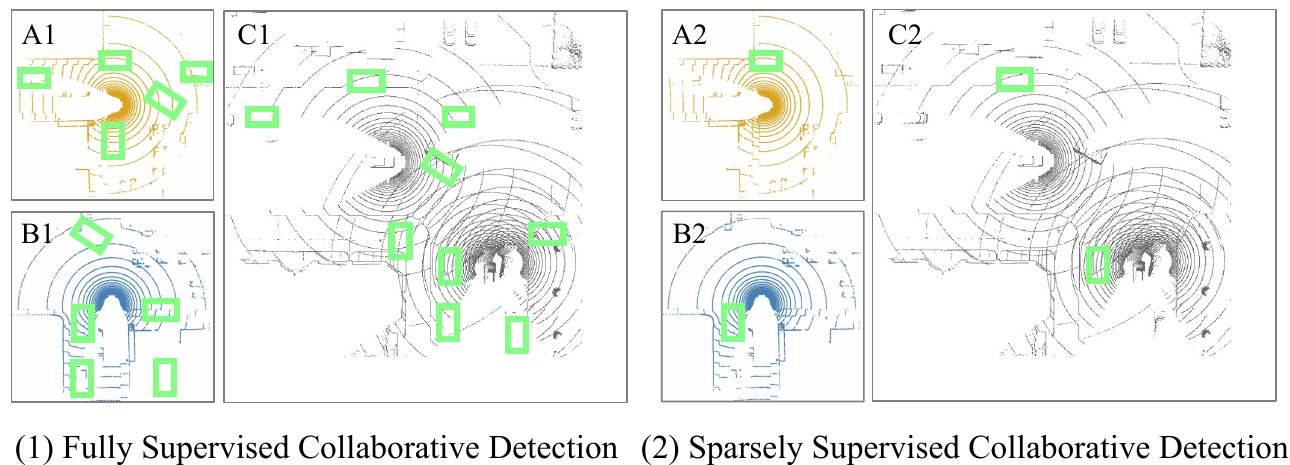}}
    \caption{Demonstration of the fully supervised and sparsely supervised collaborative 3D detection annotation. In fully supervised collaborative detection, each agent must label all targets (A1 and B1), which multiple agents merge to generate a collaborative target (C1). In contrast, sparsely supervised collaborative detection only requires each agent to randomly label one object in the scene (A2 and B2).}
    \label{fig1}
\end{figure}

While some works have explored weakly supervised 3D object detection in the single agent setting \cite{meng2020ws3d,liu2022ss3d}, none have explored this learning paradigm in collaborative 3D object detection. 
To address this research gap, we investigate collaborative 3D object detection in sparse labeling scenarios. In this setting, each agent only needs to annotate one 3D object in a scene, as illustrated on the right side of Fig. \ref{fig1}. 
This annotation strategy substantially reduces the labeling cost of the collaborative perception task, thus helping to extend it to more autonomous driving scenarios.
However, sparsely annotated 3D object detection poses new challenges as unlabeled instances can interfere with training, leading the collaborative detector to misclassify some objects as background and resulting in a significant performance decline.


To address this challenge, we propose SSC3OD, a sparsely supervised collaborative 3D object detection framework following the lightweight annotation strategy. The proposed framework aims to mitigate the adverse effects of incomplete supervision by designing two effective modules: 1) the Pillar-MAE, a masked autoencoder that pre-trains large-scale point clouds in a self-supervised manner and generates representative 3D features for unlabeled LiDAR point clouds; 2) an instance mining module that mines positive instances and generates pseudo labels for collaborative detectors online. By enhancing the 3D perception ability of the collaborative detector and mining high-quality positive instances, these two modules effectively improve the performance of sparsely supervised collaborative detectors.

To evaluate our approach, we first generate sparse labels for collaborative perception datasets in autonomous driving, including OPV2V \cite{OPV2V}, V2X-Sim \cite{V2X-Sim} and DAIR-V2X \cite{DAIR-V2X}.
In contrast to fully supervised collaborative detectors, our SSC3OD model only necessitates annotating roughly 4$\%$ of objects.
Subsequently, we perform a comprehensive evaluation of LiDAR-based collaborative 3D object detection. Both quantitative and qualitative results demonstrate that our proposed SSC3OD significantly improves the performance of sparsely supervised collaborative 3D object detection.

To summarize, our contributions are as follows:
\begin{itemize}
    \item We propose a novel framework for sparsely supervised collaborative 3D object detection from the LiDAR point clouds. To the best of our knowledge, this is the first study to investigate collaborative 3D object detection in sparse labeling scenarios.
    \item This work proposes two effective modules: Pillar-MAE, which reasons over high-level semantics in a self-supervised manner, and an instance mining module, which generates high-quality pseudo labels for collaborative detectors online.
    \item We generate sparse-labeled training sets for three large-scale collaborative perception datasets and conduct extensive experiments to evaluate the effectiveness of our proposed model. The experimental results demonstrate that our approach can significantly enhance the performance of sparsely supervised collaborative detectors.
\end{itemize}

\section{Related Work}

\subsection{Collaborative Perception}
Collaborative perception, an application of multi-agent systems \cite{han2023collaborative,yan2022pareto,gao2022performance}, has been widely applied in autonomous driving \cite{10086556,9993777,zhang2021c2fda,luo2022estnet,luo2022artificial,10059008,ye2022parallel,wang2022metaverse,wang2022dao}. During the collaborative training, agents exchange information with each other to alleviate occlusion and sensor failure. Depending on the level of transmitted data, existing works on collaborative perception can be categorized into early collaboration (raw data) \cite{cooper}, intermediate collaboration (features) \cite{fcooper, OPV2V}, and late collaboration (perception prediction) \cite{DAIR-V2X}. Besides, some datasets have been published to support research in this area, such as OPV2V \cite{OPV2V}, V2XSim \cite{V2X-Sim} and DAIR-V2X \cite{DAIR-V2X}. Several studies have explored various aspects of collaborative perception, including fusion strategies \cite{DiscoNet,CRCNet}, communication mechanisms \cite{where2comm}, localization errors \cite{coalign}, and latency issues \cite{SyncNet}. Nevertheless, there is limited research on the labeling cost of collaborative perception in autonomous driving.

\subsection{Mask Autoencoders for Point Clouds}
Masked autoencoder (MAE) \cite{he2022masked} is a straightforward self-supervised technique that learns feature representations by randomly masking patches and then reconstructing the missing pixels. Researchers have attempted to apply this approach to point clouds due to its success in 2D computer vision. Min et al. \cite{min2022voxel} propose Voxel-MAE, a masked autoencoding framework for pre-training large-scale point clouds. This framework uses a range-aware random masking strategy to mask voxels and a binary voxel classification task to learn point cloud representations. Hess et al. \cite{hess2023masked} also introduce a voxel-based masked autoencoder. They reconstruct the masked voxels and distinguish whether they are empty. 
To address the issue of insufficient data labeling in sparsely supervised learning, we propose a masked modeling method for point clouds called Pillar-MAE. Our approach randomly masks pillars and uses only the 2D encoder to learn features, which is simpler than previous MAE methods \cite{min2022voxel,hess2023masked} that rely on either Transformers or 3D encoders.

\subsection{Weakly/Sparsely Supervised 3D Object Detection}

LiDAR-based 3D detection relies on large-scale precisely-annotated data. Some recent works have proposed weakly supervised methods to reduce this heavy annotation requirement. Meng et al. \cite{meng2020ws3d} propose a two-stage weakly supervised 3D object detection framework WS3D. The first stage generates cylindrical proposals based on click-annotated bird's eye view scenes, and the second stage refines the proposals with a few well-labeled instances. Liu et al. \cite{liu2022ss3d} consider a more extreme annotation situation, sparsely supervised 3D object detection (SS3D), which only annotates one object in a scene. To address the challenge of missing annotation, they propose an instance mining module to mine positive instances and a background mining module to generate high-quality pseudo labels.
Collaborative 3D detection requires annotating objects around multiple agents in a scene, which is more time-consuming and labor-intensive than single-agent detection. Therefore, we introduce a highly weakly supervised approach where each agent only annotates one instance in the collaborative scene, called sparsely supervised collaborative 3D object detection.

\section{Methodology}
\begin{figure*}[htbp]
    \centerline{\includegraphics[width=0.8\linewidth]{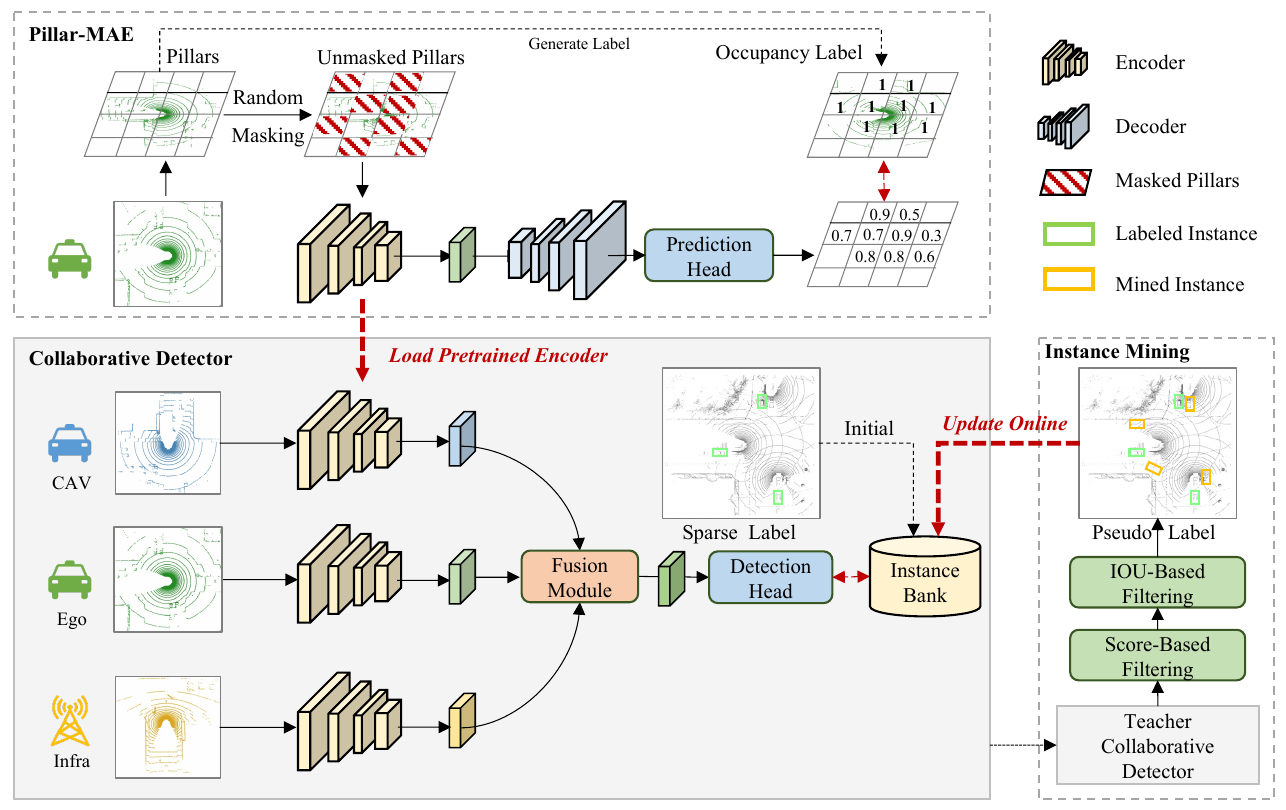}}
    \caption{The framework of SSC3OD. First, the Pillar-MAE module trains the encoder in a self-supervised manner. Second, load the pre-trained encoder and train the collaborative detector with the instance bank. The trained collaborative detector serves as the teacher collaborative detector. Third, the instance mining module identifies missing positive instances online based on the output of the teacher collaborative detector and updates the instance bank. The collaborative detectors are then retrained using the updated instance bank. (CAV refers to connected automated vehicles, and Infra refers to roadside infrastructure)}
    \label{fig2}
\end{figure*}
\subsection{Preliminary}
Consider N agents in the autonomous driving scenario, where each agent is equipped with LiDAR and can perceive objects and communicate with each other. Let $\bm{X}_i$ and $\bm{Y}_i$ be the observation and the supervision of the $i$th agent, respectively. The intermediate collaborative 3D object detection works as follows:
\begin{align}
    \label{eq1} & \bm{F}_i=f_{\rm{encoder}}(\bm{X}_i), \tag{a}\\
    & \bm{M}_{j \rightarrow i}=f_{\rm{transform}}(\xi_i,(\bm{F}_j,\xi_j)), \tag{b}\\
    & \bm{F}_{i}^{\prime}=f_{\rm{fusion}}(\bm{F}_i,\{\bm{M}_{j \rightarrow i}\}_{j=1,2,..,N}), \tag{c}\\
    \label{eq2} & \bm{O}_i=f_{\rm{head}}(\bm{F}_{i}^{\prime}), \tag{d}\\
    \label{eq3} & \bm{Y}_i^{\prime}=f_{\rm{union}}{\{\bm{Y}_i\}_{i=1,2,..,N}} , \tag{e} 
\end{align}
where $f_{\rm{encoder}}$, $f_{\rm{transform}}$, $f_{\rm{fusion}}$, $f_{\rm{head}} $ and $f_{\rm{union}}$ represent feature encoder, pose transformation module, feature fusion module, detection head and supervision union module, respectively. $\bm{F}_i$ represents the features extracted from the point cloud observed by $i$th agent, ${\xi_i=(x_i,y_i,z_i, \theta_i, \phi_i, \psi_i)}$ is the pose of the $i$th agent, $\bm{M}_{j \rightarrow i}$ refers to the $j$th agent's feature which is projected from the $j$th agent pose to the $i$th agent pose. 
After aggregating other agents' projected features, the $i$th agent's fusion feature is denoted as $\bm{F}_{i}^{\prime}$, and the detection output of the fusion feature is represented by $\bm{O}_i$.
The objective of collaborative 3D object detection is to minimize the detection loss $L_{det}$ between the perception output $\bm{O}_i$ and the collaborative supervision $\bm{Y}_i^{\prime}$, which is generated by merging the supervision of N agents as shown in step (\ref{eq3}).

It's trivial to train a collaborative detection model when each agent has complete supervision $\bm{Y}_i$. However, incomplete supervision can disrupt the model training in sparse-labeled scenarios due to the missing annotated instances. This work aims to reduce the impact of missing annotations by: 1) training the encoder in a self-supervised manner before step (\ref{eq1}) and 2) identifying missing instances after step (\ref{eq2}).

\subsection{The SSC3OD Framework}
The proposed SSC3OD is a general framework that learns robust 3D representations and mines missing positive instances for sparsely supervised collaborative detectors. As illustrated in Fig. \ref{fig2}, it comprises a collaborative detector, the Pillar-MAE module, and the instance mining module. 
The collaborative detector integrates features from multiple agents to expand the ego vehicle's field of view, and the instance bank is initialized with sparse labels, storing the collaborative detector's targets. 
Before training the collaborative detector, we use the Pillar-MAE (section \ref{A}) to pre-train the encoder in a self-supervised manner, which endows the encoder with a powerful 3D representation ability. Subsequently, we load the pre-trained encoder into the collaborative detector, which is trained with the instance bank and then acts as the teacher collaborative detector. Finally, based on the teacher collaborative detector's predictions, we mine missing instances online and merge them into the instance bank (section \ref{B}), which is used to retrain the collaborative detectors. This learning approach effectively improves the performance of the proposed sparsely supervised collaborative 3D detectors.

\subsection{Pillar-MAE Module} \label{A}
We first introduce Pillar-MAE, a masked autoencoder that pre-trains large-scale point clouds in a self-supervised manner. This approach is based on the PointPillars encoder \cite{pointpillars}.
Given the observed point clouds, Pillar-MAE randomly masks the pillars and reconstructs their occupancy values with an autoencoder network, which helps the network generate representative 3D features. After that, we detailly introduce the main components of Pillar-MAE: random masking, autoencoder, and reconstructing targets.

\textbf{Random Masking:} Given the observation $\bm{X}_i$ of the $i$th agent, the Pillar-MAE is used to generate point pillars first. For the point clouds with range $W\times H\times D$ along the $X\times Y\times Z$ axes, the pillar size is $v_W \times v_H\times D$, resulting in $n_v$ occupied pillars that contain points. We then randomly mask non-empty pillars according to the mask ratio $r_m$ and generate the occupancy label $\bm{T}$ of the reconstruction task. In the occupancy label, the value of occupied and empty pillars are 1 and 0, respectively.

\textbf{Autoencoder:} The autoencoder consists of a 2D encoder and a 2D decoder. Following PointPillars, we transform the unmasked pillars into pseudo images and use a 2D convolutional encoder to extract features. Our decoder consists of a lightweight 2D deconvolution layer, which transforms the encoded features to the original size of the pillars.

\textbf{Reconstructing Target:} Based on decoded features, we predict occupied pillars $P$ and adopt a simple binary cross entropy loss as occupancy loss $L_{occ}$:
\begin{equation}
    L_{occ}=-\frac{1}{b} \sum_{i=1}^{b} \sum_{j=1}^{n_l} \bm{T}_j^i \log \bm{P}_j^i
\end{equation}
where $b$ is the batch size, $\bm{P}_j^i$ is the probability of $j$th pillar of the $i$th training sample, and $\bm{T}_j^i$ is the corresponding ground truth whether the pillar contains point cloud.

With Pillar-MAE, the encoder is compelled to acquire high-level features for reconstructing the masked occupancy distribution of the 3D scene using only a small number of visible pillars. Subsequently, we utilize the pre-trained encoder to initialize and fine-tune the collaborative detectors with the sparse-labeled dataset.

\subsection{Instance Mining Module}\label{B}
Although the pre-trained encoder has strong representation ability, the detector still has difficulty in distinguishing some foreground instances from the background due to the lack of complete supervision. To further enhance the perceptual ability of the detector, we design an instance mining module to mine unlabeled instances. Specifically, we utilize the trained detector as a teacher collaborative detector to guide the instance mining module in identifying missing instances.

The instance mining module employs both score-based and IoU-based filtering mechanisms. The score-based filtering selects high-confidence predictions, while the IoU-based filtering leverages Non-Maximum Suppression (NMS) to eliminate overlapping detections. To ensure the quality of the mined instances, we set a relatively high score threshold $\tau_{cls}$ and a relatively low IoU threshold $\tau_{IOU}$. The high-quality pseudo labels can be generated by applying these two filtering mechanisms.

To emulate real-world scenarios, we randomly select the ego vehicle in the training stage of collaborative detection, which results in varying input and detection targets in each epoch. As it is not feasible to save mined instances offline in this situation, built upon knowledge distillation, we employ a teacher collaborative detector to guide the learning of collaborative detectors. The process involves mining instances online and merging them into the instance bank. Algorithm \ref{alg1} outlines the pipeline of the instance mining module.

\begin{algorithm}
    \caption{Instance Mining Module}
    \label{alg1} 
    \renewcommand{\algorithmicrequire}{\textbf{Input:}}
    \renewcommand{\algorithmicensure}{\textbf{Output:}}
    \begin{algorithmic}[1]
        \Require Pre-trained encoder $\bm{M}$, collaborative detector $\bm{F}$, teacher collaborative detector $\bm{F}_t$, point clouds $\bm{X}$, sparse label $\bm{\hat{Y}}$, instance bank $\bm{B}$, score threshold $\tau_{cls}$, IOU threshold $\tau_{IOU}$, training epoch $E$.
        \State $\bm{B} \leftarrow \bm{\hat{Y}}$ \Comment{Initialize instance bank}
        \State $\bm{F}_1$ = $Load(\bm{F}_1\leftarrow\bm{M}$) \Comment{Load pre-trained encoder}
        \For {$e = 1,2,...,E$}
        \For {$x$ in $\bm{X}$}
        \State $\bm{O}=\bm{F}_1(x)$
        \State Calculate Detection Loss $L_{det}$ between $\bm{O}$ and $\bm{B}$
        \State Update the weight of $\bm{F}_1$ by minimizing $L_{det}$
        \EndFor
        \EndFor
        \State $\bm{F}_t\leftarrow\bm{F}_1$ \Comment{Trained detector as teacher detector}
        \State $\bm{F}_2$ = $Load(\bm{F}_2\leftarrow\bm{M}$) \Comment{Load pre-trained encoder}
        \For {$e = 1,2,...,E$}
        \For {$x$ in $\bm{X}$}
        \State $\bm{O}_t=\bm{F}_t(x)$
        \State Select $\hat{\bm{O}}_t$ if $\hat{\bm{O}}^{cls}_t>\tau_{cls}$\Comment{Score-based filtering}
        \State Select $\widetilde{\bm{O}}_t$ with NMS($\tau_{IOU})$\Comment{IOU-based filtering}
        \State $\bm{B} \leftarrow \bm{B} + \widetilde{\bm{O}}_t$ \Comment{Update instance bank}
        \State $\bm{O}=\bm{F}_2(x)$
        \State Calculate Detection Loss $L_{det}$ between $\bm{O}$ and $\bm{B}$
        \State Update the weight of $\bm{F}_2$ by minimizing $L_{det}$
        \EndFor
        \EndFor
        \Ensure Collaborative detector $\bm{F}_2$
    \end{algorithmic}
\end{algorithm}



\section{Experiments}
We conduct experiments on three large-scale collaborative perception datasets consisting of both real-world and simulated scenarios involving two types of agents. We choose three types of intermediate collaboration methods, including traditional \cite{fcooper}, graph-based \cite{DiscoNet} and attention-based \cite{OPV2V} fusion methods.
The input considered is solely LiDAR, and we assume an ideal collaborative perception scenario without latency or pose error.
The LiDAR-based collaborative 3D object detection performance is measured with Average Precisions (AP) at Intersection-over-Union (IoU) thresholds of 0.3, 0.5 and 0.7. 

\subsection{Datasets}
\textbf{DAIR-V2X} \cite{DAIR-V2X} is the first real-world vehicle-to-infrastructure (V2I) collaborative perception dataset, which contains a vehicle and a roadside unit in each cooperative frame. We adopt the complete cooperative annotation from CoAlign \cite{coalign} and set the LiDAR range as $x\in[-100m,100m],y\in[-40m,40m]$.

\textbf{OPV2V} \cite{OPV2V} is a simulated vehicle-to-vehicle (V2V) collaborative perception dataset, which is collected with the co-simulating framework OpenCDA \cite{opencda} and CARLA simulator \cite{carla}. It contains one to five vehicles in each cooperative frame, and we set the LiDAR detection range as $x \in [-140m,140m],y \in [-40m,40m]$.

\textbf{V2X-Sim} \cite{V2X-Sim} is a simulated vehicle-to-everything (V2X) collaborative perception dataset. It is generated with traffic simulation SUMO \cite{sumo} and CARLA simulator \cite{carla}, including 100 scenes with 10,000 frames divided into 8,000/1,000/1,000 for training/validation/testing. We use V2XSim 2.0 and set the LiDAR range as $x\in[-32m, 32m], y\in[-32m,32m]$ for collaborative 3D detection.

We expand the sparse labeling of the original collaborative perception dataset by randomly retaining a single annotated object of each agent in every 3D scene from the training set. As shown in Tab. \ref{table:label number}, we count the number of frames in training sets for three datasets, along with the number of both full and sparse labels, and compute the proportion of sparse labels to all objects. The sparse subsets require annotation of only around $4\%$ of objects, in contrast to the complete annotation of all objects in the original training set. This demonstrates the cost-saving effect of the proposed sparse annotation strategy.

\begin{table}[]
    \renewcommand\arraystretch{1.3} 
    \centering 
    \caption{Comparison of full and sparse annotation in collaborative detection.}
    \resizebox{\linewidth}{!}{
    \begin{tabular}{cc|ccc}
    \toprule 
    Datasets & \# Train size &\# Full label & \# Sparse label & Sparse ratio \\
    \midrule
    DAIR-V2X \cite{DAIR-V2X} &4811 & 237725 &	9622	&$\textbf{4.05\%}$ \\
    OPV2V \cite{OPV2V} &6765&358142&21139&$\textbf{5.90\%}$ \\
    V2X-Sim \cite{V2X-Sim}& 8000& 698991 &	29300	& $\textbf{4.19\%}$ \\
    \bottomrule
    \end{tabular}
    }
    \label{table:label number}
\end{table}

\subsection{Implementation Details}
We adopt PointPillars \cite{pointpillars} with the grid size of $[0.4m,0.4m]$ as our backbone. During training, we randomly select an autonomous vehicle (AV) as the ego vehicle, whereas a fixed ego vehicle is employed for detector evaluation during testing. 
We adopt single-scale intermediate fusion and training with Adam optimization. The training epoch of Pillar-MAE is 25, and the training epoch of collaborative detectors are 20, 30, and 20 on DAIR-V2X, OPV2V, and V2X-Sim, respectively. We set the mask ratio $r_m$ in Pillar-MAE as 0.7. For instance mining module, we set the score threshold $\tau_{cls}$ and IOU threshold $\tau_{IOU}$ as 0.3 and 0.15, respectively. All models are trained on RTX A4000. 

We choose three novel intermediate collaborative detection methods: F-Cooper \cite{fcooper}, AttFusion \cite{OPV2V} and DiscoNet \cite{DiscoNet}. F-Cooper employs a straightforward maxout method to fuse features, whereas AttFusion introduces a single-head self-attention fusion module. DiscoNet combines matrix-edge valued weight with early collaboration based knowledge distillation to capture feature relationships.
To ensure a fair and impartial comparison, we only employ the graph fusion module from DiscoNet, omitting the early collaborative knowledge distillation component.

\subsection{Quantitative Results}
To validate the performance of the proposed framework, we train the collaborative detectors with three distinct strategies, 1) training from scratch on the full-labeled training set, 2) training from scratch on the sparse-labeled training set, and 3) training with the SSC3OD framework on the sparse-labeled training set. Tab. \ref{table:performance} shows the performance of detectors trained with three strategies. 
Compared with the fully supervised collaborative detector, the performance of the sparsely supervised collaborative detector drops more than 10$\%$ when trained from scratch. This decline can be attributed to most positive instances remaining unlabeled in sparsely supervised scenarios, causing the model to identify them as background. However, training the sparsely supervised collaborative detector with our SSC3OD framework (w/ SSC3OD) yields significantly improved performance, indicating the framework's effectiveness in mitigating incomplete annotation impact.

\begin{table*}[htbp]
    \renewcommand\arraystretch{1.3} 
    \centering 
    \caption{Comparison of collaborative detectors trained with the full-labeled and sparse-labeled datasets.}
    \resizebox{\linewidth}{!}{
    \begin{tabular}{c|c|ccc|ccc|ccc}
    \toprule 
    \multicolumn{2}{c}{Datasets}&\multicolumn{3}{c}{DAIR-V2X \cite{DAIR-V2X}}&\multicolumn{3}{c}{OPV2V \cite{OPV2V}}&\multicolumn{3}{c}{V2X-Sim \cite{V2X-Sim}}\\
    \midrule
    Methods& Settings& AP@0.3 &AP@0.5 &AP@0.7& AP@0.3 &AP@0.5 &AP@0.7& AP@0.3 &AP@0.5 &AP@0.7\\
    \midrule
    1. F-Cooper \cite{fcooper} & Full& 80.11 & 73.37 & 56.51& 94.23& 88.44& 69.16 & 79.69 & 71.59 & 56.75\\
    2. F-Cooper \cite{fcooper} & Sparse& 54.41 & 49.49 & 32.58 & 80.83&74.75&51.96 & 62.62 & 52.14 & 38.72\\
    3. F-Cooper (w/ SSC3OD) & Sparse & 60.89 & 56.68 & 40.53& 84.40& 79.08 & 57.54 &  71.20  &  62.49 &  50.09 \\
    4. Improvements 2$\rightarrow$3& - &\textbf{+6.48} & \textbf{+7.19}&\textbf{+7.95}& \textbf{+3.57}&\textbf{+4.33}&\textbf{+5.58} & \textbf{+8.58}&\textbf{+10.35} &\textbf{+11.37}\\
    \midrule
    \midrule
    1. AttFusion \cite{OPV2V} & Full& 79.82 & 72.89 & 57.43 &95.55&93.49&83.03&79.07 & 76.47 & 66.43\\
    2. AttFusion \cite{OPV2V} & Sparse& 56.31 & 51.21 & 32.93&78.74&77.63&61.98&67.02 & 65.53 & 56.53\\
    3. AttFusion (w/ SSC3OD) & Sparse & 62.02 & 57.17& 40.57 & 82.35 & 81.64 & 72.55 & 74.10  & 72.68 & 62.66\\
    4. Improvements 2$\rightarrow$3& - &\textbf{+5.71} &\textbf{+5.96} &\textbf{+7.64}& \textbf{+3.61}&\textbf{+4.01}&\textbf{+10.57}& \textbf{+7.08}&\textbf{+7.15}&\textbf{+6.13}\\
    \midrule
    \midrule
    1. DiscoNet \cite{DiscoNet}& Full& 80.21 & 73.62 & 58.00& 96.22&94.31 & 84.42 & 81.02 & 77.63 & 69.29\\
    2. DiscoNet \cite{DiscoNet}& Sparse& 55.90  & 51.39 & 32.97&81.82&80.79&69.23&68.56 & 65.39 & 55.84\\
    3. DiscoNet (w/ SSC3OD) & Sparse &  60.93 &  56.78 &  42.63 & 83.12 & 82.22 & 73.56& 74.07 &  70.54 &62.29\\
    4. Improvements 2$\rightarrow$3& - &\textbf{+5.03} &\textbf{+5.39} &\textbf{+9.66}&\textbf{+1.30}&\textbf{+1.43}&\textbf{+4.33}& \textbf{+5.51} & \textbf{+5.15} & \textbf{+6.45} \\
    \bottomrule
    \end{tabular}
    }
    \label{table:performance}
\end{table*}

\begin{figure*}[htbp]
    \centerline{\includegraphics[width=0.9\textwidth,height=0.4\textheight]{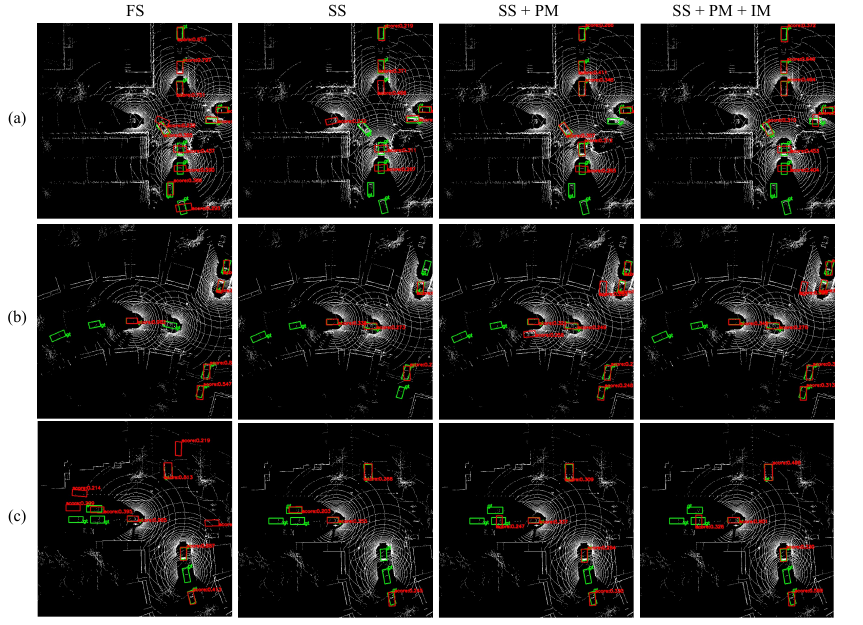}}
    \caption{Qualitative comparison of collaborative detectors with different settings in V2X-Sim. Green boxes are ground-truth and red ones are detection outputs. We choose the detection results of (a) F-Cooper, (b) AttFusion and (c) DiscoNet to show the effect of modules in SSC3OD. Zoom in to view more details.}
    \label{fig3}
\end{figure*}

\subsection{Ablation Studies}
This section presents a series of ablation studies to analyze modules' contributions in SSC3OD. 
Tab. \ref{table:ablation study} displays the performance of collaborative detectors using various settings on sparse-labeled V2XSim. We observe that: i) directly utilizing the instance mining module (IM) on the collaborative detector does not significantly improve performance due to the unreliable results of sparsely supervised collaborative detectors trained from scratch. ii) The performance of the collaborative detector loaded with the pre-trained encoder has been significantly improved, thereby demonstrating the effectiveness of the Pillar-MAE module (PM) in enhancing the point cloud perception ability of the encoder. iii) Using IM after a PM-based collaborative detector leads to further improvements in the performance of collaborative detectors, revealing that the PM-based collaborative detector provides reliable detection results and IM mines high-quality pseudo labels.

Then, we analyze the impact of the mask ratio $r_m$ in Pillar-MAE. We train the Pillar-MAE with different $r_m$ values (0.5, 0.7 and 0.9) and then utilize corresponding pre-trained encoders to train collaborative detectors. Tab. \ref{table:mask ratio} illustrates the effect of the mask ratio in Pillar-MAE. We observe that the PM-based collaborative detector achieves optimal performance when $r_m=0.7$. This is because the model under the small $r_m$ retains more point cloud information, and the occupancy of the point cloud can be reconstructed without too much perception ability. Conversely, a large $r_m$ leads to excessive loss of point cloud information, compromising effective point cloud perception.

In addition, we analyze the impact of the score threshold $\tau_{cls}$ in the instance mining module. We regard the collaborative detector trained with the pre-trained encoder as the teacher collaborative detector and set different values of $\tau_{cls}$ to generate pseudo labels. As the classification score of the sparsely supervised collaborative detector falls between 0.2 and 0.35, we set $\tau_{cls}$ to 0.2, 0.25 and 0.3. Tab. \ref{table:score threshold} demonstrates a consistent enhancement in detection performance when $\tau_{cls}$ is set to 0.3. When $\tau_{cls}$ is set to 0.2 or 0.25, the pseudo labels may contain more false positive instances. Therefore, we choose $\tau_{cls}=0.3$ to ensure high-quality pseudo labels.

\begin{table}[htbp]
    \renewcommand\arraystretch{1.2} 
    \centering 
    \caption{Component ablation studies on sparse-labeled V2X-Sim dataset.}
    \resizebox{\linewidth}{!}{
    \begin{tabular}{cc|ccc}
    \toprule 
    \multicolumn{2}{c}{Settings} &\multicolumn{3}{c}{AP@0.50 / AP@0.70}  \\
    \midrule
    PM & IM &F-Cooper& AttFusion  & DiscoNet \\
    \midrule
    -&-& 52.14 / 38.72 &	65.53 / 56.53	& 65.39 / 55.84\\
    \midrule
    -&\checkmark&54.36 / 41.06& 67.67 / 56.89& 65.37 / 55.05 \\
    \checkmark&-&60.31 / 45.60&68.41 / 59.06&68.21 / 59.06 \\
    \checkmark&\checkmark& 62.49 / 50.09 & 72.68 / 62.66 & 70.54 / 62.29 \\
    \bottomrule
    \end{tabular}
    }
    \begin{tablenotes}
        \footnotesize
        \item PM refers to Pillar-MAE, IM refers to instance mining.
    \end{tablenotes}
    \label{table:ablation study}
\end{table}

\begin{table}[htbp]
    \renewcommand\arraystretch{1.2} 
    \centering 
    \caption{Ablation studies of mask ratio in Pillar-MAE module on sparse- labeled V2X-Sim dataset.}
    \resizebox{\linewidth}{!}{
    \begin{tabular}{c|ccc}
    \toprule 
    \multicolumn{1}{c}{Settings} &\multicolumn{3}{c}{AP@0.50 / AP@0.70}  \\
    \midrule
    $r_m$ &F-Cooper& AttFusion  & DiscoNet\\
    \midrule
    -& 52.14 / 38.72 &	65.53 / 56.53	& 65.39 / 55.84\\
    \midrule
    0.5& 59.26 / 45.92& 68.46 / 58.21& 66.57 / 54.64\\
    0.7& 60.31 / 45.60& 68.41 / 59.06 &68.21 / 59.06 \\
    0.9& 55.29 / 43.66 & 63.47 / 52.67 & 68.07 / 58.41 \\
    \bottomrule
    \end{tabular}
    }
    \begin{tablenotes}
        \footnotesize
        \item - refers to training from scratch.
    \end{tablenotes}
    \label{table:mask ratio}
\end{table}

\begin{table}[htbp]
    \renewcommand\arraystretch{1.2} 
    \centering 
    \caption{Ablation studies of score threshold in instance mining module on sparse-labeled V2X-Sim dataset.}
    \resizebox{\linewidth}{!}{
    \begin{tabular}{c|ccc}
    \toprule 
    \multicolumn{1}{c}{Settings} &\multicolumn{3}{c}{AP@0.50 / AP@0.70}  \\
    \midrule
    $\tau_{cls}$ &F-Cooper& AttFusion  & DiscoNet \\
    \midrule
    -& 52.14 / 38.72 &	65.53 / 56.53	& 65.39 / 55.84\\
    w/ PM& 60.31 / 45.60&68.41 / 59.06&68.21 / 59.06\\
    \midrule
    0.20& 65.94 / 51.03& 69.95 / 54.92& 73.81 / 61.33\\
    0.25& 63.87 / 49.21& 72.03 / 58.65& 72.02 / 62.78 \\
    0.30& 62.49 / 50.09 & 72.68 / 62.66 & 70.54 / 62.29 \\
    \bottomrule
    \end{tabular}
    }
    \label{table:score threshold}
    \begin{tablenotes}
        \footnotesize
        \item - refers to training from scratch, w/ PM refers to training with PM.
    \end{tablenotes}
\end{table}

\subsection{Qualitative Results}
This section presents a qualitative evaluation of the proposed module. We show the detection results of the fully supervised collaborative detectors (FS), sparsely supervised collaborative detectors trained from scratch (SS), sparsely supervised collaborative detectors trained with Pillar-MAE (SS + PM), and sparsely supervised collaborative detectors trained with Pillar-MAE and instance mining module (SS + PM + IM).

Fig. \ref{fig3} illustrates the collaborative detection results in different settings. Detectors trained in the FS setting achieve high-confidence detection results. Conversely, detectors trained in the SS setting exhibit relatively low-confidence detection results due to missing positive instance labels. Collaborative detectors trained in the SS + PM setting recognize more foreground and reduce false positives, indicating the encoder's powerful 3D perception ability. Moreover, adding the instance mining module to detectors (SS + PM + IM) enhances object localization accuracy and substantially increases confidence scores. This demonstrates the effectiveness of the instance mining module in producing high-quality pseudo labels.

\section{Conclusion}
This work introduces SSC3OD, a sparsely supervised collaborative 3D object detection framework. The proposed framework comprises two key modules: the Pillar-MAE and the instance mining module. The Pillar-MAE enhances the 3D perception ability of encoders in a self-supervised manner, while the instance mining module generates high-quality pseudo labels for collaborative detectors online. The sparse labels for collaborative perception datasets are generated to evaluate the proposed framework. Extensive experiments demonstrate that SSC3OD significantly improves the performance of sparsely supervised collaborative 3D object detection.

\bibliographystyle{IEEEtran}

\bibliography{reference}

\end{document}